\documentclass{ifacconf}

\usepackage{graphicx}      
\usepackage{subfigure}

\usepackage{amsmath,amsfonts,amssymb}
\usepackage{multirow}
\usepackage{natbib}        
\usepackage[colorinlistoftodos]{todonotes}
\usepackage{footnote}
\usepackage{float}
\usepackage{balance}
\begin{document}
\begin{frontmatter}

\title{Computationally efficient dense moving object detection based on reduced space disparity estimation}

\thanks[footnoteinfo]{This work has been supported by the European Union’s Horizon 2020 research and innovation programme under grant agreement No. 688117 (SafeLog) and by the Croatian Science Foundation under contract No. I-2406-2016. This work has also been supported by the FLAG-ERA JTC 2016 and the Ministry of Science and Education of the Republic of Croatia under the project Rethinking Robotics for the Robot Companion of the future (RoboCom++).}

\author[]{Goran Popovi\'c,}
\author[]{Antea Hadviger,}
\author[]{Ivan Markovi\'c,}
\author[First]{Ivan Petrovi\'c}

\address[First]{University of Zagreb Faculty of Electrical Engineering and Computing, Laboratory for Autonomous Systems and Mobile Robotics, Zagreb, Croatia
    (e-mail: name.surname@fer.hr).}

\begin{abstract}                
Computationally efficient moving object detection and depth estimation from a stereo camera is an extremely useful tool for many computer vision applications, including robotics and autonomous driving.
In this paper we show how moving objects can be densely detected by estimating disparity using an algorithm that improves complexity and accuracy of stereo matching by relying on information from previous frames.
The main idea behind this approach is that by using the ego-motion estimation and the disparity map of the previous frame, we can set a prior base that enables us to reduce the complexity of the current frame disparity estimation, subsequently also detecting moving objects in the scene.
For each pixel we run a Kalman filter that recursively fuses the disparity prediction and reduced space semi-global matching (SGM) measurements.
The proposed algorithm has been implemented and optimized using streaming single instruction multiple data instruction set and multi-threading.
Furthermore, in order to estimate the process and measurement noise as reliably as possible, we conduct extensive experiments on the KITTI suite using the ground truth obtained by the 3D laser range sensor.
Concerning disparity estimation, compared to the OpenCV SGM implementation, the proposed method yields improvement on the KITTI dataset sequences in terms of both speed and accuracy.
\end{abstract}


\end{frontmatter}

\section{Introduction}

Depth estimation plays an important role in various autonomous mobile robot environment inference applications; such as simultaneous localization and mapping, navigation\,---\,particularly obstacle detection and avoidance.
Even though depth can be estimated using a variety of sensors, cameras are still a very popular choice due to their low price, high availability, and low weight, especially when payload is a critical factor as in small unmanned aerial vehicles.
The depth, or disparity maps (DMs), can be computed from the intensity images obtained by either monocular or stereo cameras.
In the analytical domain, general categorization splits the disparity computation methods into two groups: local methods and global methods.
Local methods are faster and prone to discontinuities as they consider only a neighborhood of a pixel to determine its depth.
On the other hand, global methods typically provide continuous DMs due to optimizing a global cost function, including the smoothness terms, making the computation more complex.
Another class of currently promising approaches are deep-learning based methods, where the DM estimation is learned from a sequence of animated images with readily available ground truth.
However, in the present paper we focus on the computational efficiency of an analytical approach, and do not discuss further learning based methods in the present paper.

\cite{Hirschmuller2008} introduced an approach that reduces the complexity of global methods, but still keeps the smoothness of the obtained DM.
The proposed method, dubbed semi-global matching (SGM), approximates the global cost function by recursively aggregating costs in several directions across the image, making it faster to compute than global methods, as well as discontinuity aware.
SGM is one of the state-of-the-art algorithms for DM estimation and still nowadays researchers try to modify it in order to make it faster and more accurate. 
One approach is to optimize the hardware architecture for fast disparity computation,  \cite{Li2017} and \cite{Hofmann2016}.
On the other hand, faster execution can be achieved by reducing algorithm's complexity or by executing more tasks simultaneously.
Most common approaches to complexity reductions are achieved by reducing the image resolution \citep{Gehrig2010}, reducing the number of accumulation paths \citep{Hermann2011}, and reducing the disparity range \citep{Fucek2017}. 
Apart from reducing the algorithm's complexity, execution time can also be improved with studious implementation.
\cite{Spangenberg2014} proposed exploiting the CPU architecture by using SIMD instructions and multi-threading.
Therein the authors modified the original SGM algorithm by compressing the disparity space (subsampling) and separating the image into horizontal stripes to enable multi-threading in the step of cost accumulation.
Newer approaches improve the compuation accuracy with neural networks as in \cite{Seki2017} where the authors use learned penalties for disparity jumps.

%

Since DM computation is often computed on a sequence of images, it is natural to use the information computed in  previous steps to help improve computation in the following ones, in terms of both speed and accuracy.
Utilization of recursive filters, such as the Kalman filter, has been described in the works of \cite{Matthies1989}, \cite{Agrawal2005}, \cite{Morales2013}.
Work of \cite{Matthies1989} introduces a pixel based approach meaning that Kalman filtering is performed on all the pixels in the image, and not just on distinct points set.
In \cite{Morales2013} authors integrated and compared several DM computation algorithms within a Kalman filtering framework.
One of the compared DM algorithms was SGM, which served as a measurement generator for the Kalman filter.
Since Kalman filtering yields prediction of the system's state, consequently it also offers statistical ground for detecting moving objects in dynamic scenes.
In the work of \cite{Agrawal2005} authors used a similar approach to track a moving person.
Therein, the estimated motion of the camera was used to predict the disparity image, which was then compared with the actual measurement and the areas with low similarity value were grouped into blobs.
Blobs were then tracked with the Kalman filter, thus tracking the moving people.

In the present paper we draw upon our earlier work of \cite{Fucek2017} and we fuse the discussed ideas and approaches, in order to produce a computationally efficient algorithm capable of detecting moving object using the reduced disparity search space.
We use pixel-wise Kalman filters to estimate the disparity on reduced search space and densely detect parts of the image which contain moving objects.
Thereafter, pixels are grouped in order to get a single and more robust detection indicating moving objects.
The implemented algorithm is evaluated on the KITTI dataset \citep{Geiger2013} and compared with OpenCV's implementation of SGM in terms of execution speed and accuracy.
Furthermore, we use the disparity ground truth from the 3D laser range sensor of the KITTI sequences, to estimate the process noise, i.e., the uncertainty of the ego-motion, and measurement noise, i.e., the uncertainty of the SGM-based disparity estimation.

The rest of the paper is organized as follows.
First, we describe the steps of the reduced search space SGM algorithm and its efficient implementation in Section~\ref{sec:reduced_sgm}.
In Section~\ref{sec:q_and_r_estimation} we describe how we estimate the process and measurement noise of the Kalman filter and in Section ~\ref{sec:mov_obj_detection} we show how the enhancement of the algorithm can detect the movement of objects in the scene.
Experimental evaluation and comparison with the OpenCV implementation of SGM is given in Section~\ref{sec:evaluation}, while Section~\ref{sec:conclusion} concludes the paper.

\section{Reduced space Semi-Global Matching}\label{sec:reduced_sgm}

We have modified the general framework of SGM in order to enable reduction of the disparity search space.
Apart from the intensity image pair, SGM also inputs two values for each pixel that define the range for disparity search.
The algorithm has been studiously implemented for efficiency using SSE instruction set and multi-threading.
In the sequel we describe the implemented algorithm in details.

\subsection{Matching cost calculation}

The first step of all stereo matching algorithms is the computation of the matching cost that measures the similarity between the base image patches and potentially matching ones in the matching image.
Since the KITTI dataset consists of outdoor environment images with poorly controlled lightning, it is necessary to choose a matching cost method invariant to changes in brightness.
The implemented algorithm uses the census-based matching cost, proven to perform well in the described conditions \citep{Hirschmuller2009}.
Census transform describes a pixel using the relative intensity of its neighboring pixels inside an $n \times m$ window.
The final cost of matching a pair of pixels is calculated as the Hamming distance between the two obtained census transforms.

For real-time applications, window size of $5 \times 5$ pixels yields good performance in terms of both speed and accuracy \citep{Gehrig2010}.
The matching cost computation step is performed $height \cdot width \cdot number\ of\ disparities$ times per frame, making it one of the most time-consuming parts of the algorithm.
Census transforms of all the pixels are calculated using the SSE instruction set, allowing processing of 16 pixels of an 8 bit image at once using the 128-bit SEE registers.
However, due to additional instructions necessary to correctly align the data, the speedup is lower than 16 times. 
The final matching cost is calculated using the \texttt{xor} operation supported by SSE, followed by a population count.
Both census transform and cost computation are
parallelized by using multiple threads for corresponding horizontal stripes of the image as the calculation is completely independent for each pixel.

\subsection{Cost aggregation}

Semi-global matching propagates information throughout the image by recursively aggregating matching costs and discontinuity penalties along several one-dimensional paths using the following equation:

\begin{equation} \label{eq:loss_formula}
	L_r(\textbf{p},d)=C(\textbf{p},d)+\min
			\begin{cases}
				L_r(\textbf{p}-\textbf{r},d)\\
				L_r(\textbf{p}-\textbf{r},d \pm 1) +P_1\\
				\min_{i} L_r(\textbf{p}-\textbf{r},d + i) +P_2\\
			\end{cases}
\end{equation}

\begin{figure*}[!t]
  \centering
	\includegraphics[width=0.68\textwidth]{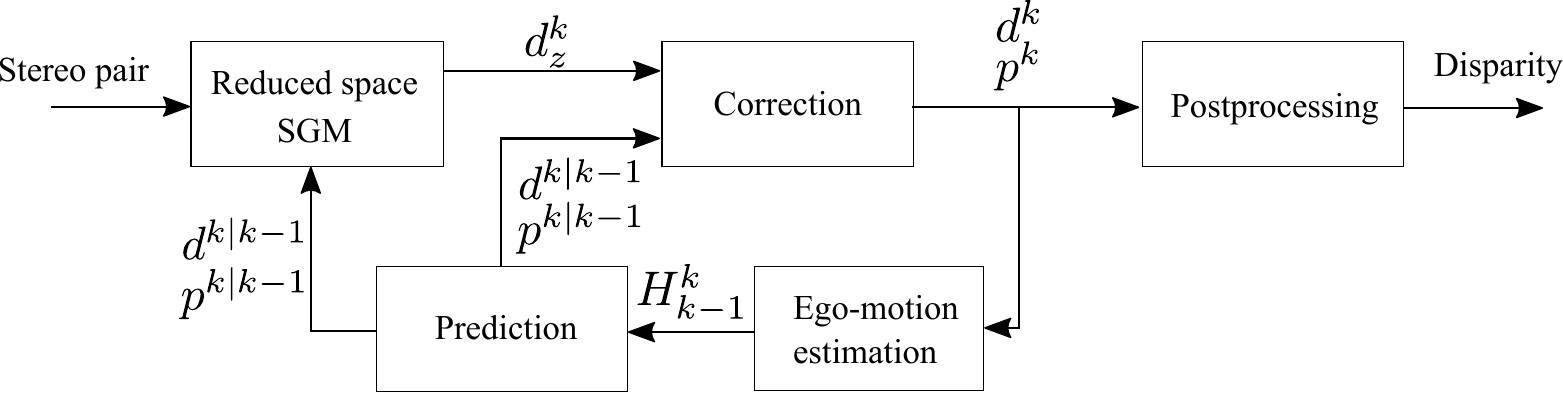}
	\caption{Block diagram of proposed SGM algorithm.}
	\label{fig:diagram}
\end{figure*}

Path accumulation is the most time-critical part of the algorithm, its complexity being $\mathcal{O}(number\ of\ paths \cdot height \cdot width \cdot number\ of\ disparities)$.
Usually, the number of individual paths is 4, 8 or 16, which directly affects execution time.
In the conducted experiments, presented in Section~\ref{sec:evaluation}, we have noticed that increasing the number of paths above 4 did not justify the increased execution time, but on the contrary yielded similar results in accuracy.
The implementation itself is done using AVX2 instructions, operating on 256-bit vectors.
Since accumulated loss can fit into 16-bit integers, it is possible to process 16 disparity values at a time.
Costs of individual paths are aggregated into the final loss in the first pass going from the top left to the bottom right, and in the second pass going from the bottom right to the top left.
Since this information needs to be propagated throughout the image, it is not possible to parallelize this step on the thread level as in the case of matching cost calculation.

\subsection{Disparity computation and refinement}

Disparity values are obtained using the simple \textit{winner-takes-all} approach, meaning that the chosen disparity for each pixel is the one with the minimum accumulated loss across all paths.
Parallelization in this step is done using SIMD instructions and by dividing the image into horizontal stripes on thread level.
Despite the smoothing constraints, noise can be present in the final disparity map.
In postprocessing, a block median filter of $3 \times 3$ pixels is used to reduce noise, fill the values of possibly invalidated disparities, and additionally improve smoothness.
It is worth noting that the median filter is outside of the Kalman loop (Figure~\ref{fig:diagram}), thus not affecting the recursive filtering process.
However, the downside of median blurring is that it can negatively affect original correctly calculated disparity values.

\subsection{Disparity estimation via Kalman filtering}

As previously described in the work of \cite{Fucek2017}, Kalman filter is used to estimate the disparity and its variance.
For completeness, herein we provide the equations of the filtering recursion.
The prediction step estimates the coordinates and disparity of all pixels from the frame $k-1$ in the next frame $k$ based on the camera ego-motion.
The filter is applied directly in the disparity state-space as suggested by \cite{Demirdjian2001}. 
In such a state-space representation, the predicton step of the Kalman filter is given by the following equation:
	\begin{equation} \label{eq:prediction}
		\begin{bmatrix} \omega^{k|k-1} \\ 1 \end{bmatrix} = H^{k}_{k-1} \begin{bmatrix} \omega^{k-1} \\ 1	\end{bmatrix},
	\end{equation}
where $\omega = \begin{bmatrix} x & y & d \end{bmatrix}^{\top}$ are the coordinates in the disparity space and $H^{k}_{k-1}$ is the $4\times 4$ transformation matrix.

For the computation of disparity variance, we used a motion model presented in \cite{Demirdjian2001} where the model is defined as the following ratio of the predicted and current disparity:
	\begin{equation} \label{eq:prediction_model}
		\Phi^{k-1}=d^{k|k-1} / d^{k-1}.
	\end{equation}
The variance of the prediction is then computed as:
	\begin{equation} \label{eq:prediction_var}
		p^{k|k-1}=\left(\Phi^{k-1} \right)^2 p^{k-1} + q^{k-1},
	\end{equation}
where $q^{k-1}$ is the process noise.

Incorrect disparity predictions are expected near object edges, i.e. disparity discontinuities, due to possible errors in ego-motion estimation. To allow correction of the introduced errors in the stereo matching phase, we reject all the predicted disparities near disparity discontinuities and reset their variance to cover the whole disparity space.

Camera's forward movements introduce holes in the disparity prediction which is known as the \textit{zooming effect}.
To cancel it out, we define a new value of disparity and variance for each pixel that has either 2 vertical or 2 horizontal neighbors with known disparity.
In that case pixel's disparity and variance are defined as the average of its neighbors.
Although in (\ref{eq:prediction}) the vector $\omega$ has three variables, the steps of Kalman filter are computed for the disparity only, while for the first two coordinates we do not consider their statistical properties.

After the prediction step, the predicted disparity mean and variance are forwarded to SGM to set the limits on the search space to ($d^{k|k-1}\pm p^{k|k-1}$) in the cost aggregation step, corresponding to the possible values of $i$ in the equation (\ref{eq:loss_formula}).
Therefore, the search space is reduced proportionally to the uncertainty of the predicted disparity.
Given the measured disparity that is then obtained from SGM, it is possible to perform the correction step:
	\begin{equation} \label{eq:correction_mean}
		d^{k}=d^{k|k-1} + K_k \left( d^k_z - d^{k|k-1} \right)
	\end{equation}
	\begin{equation} \label{eq:correction_variance}
		p^{k}= \left( 1-K_k \right)^2    p^{k|k-1} + K_k^2r^k
	\end{equation}
where $K_k$ is the Kalman gain.

Figure~\ref{fig:diagram} shows the block diagram of the overall algorithm.
Parameters $d^{k|k-1}$ and $p^{k|k-1}$, which define the search limits, are initially set to cover the whole search space, effectively turning our approach into the conventional SGM.
It should also be pointed out that the current algorithm can also work with other recursive estimators that can provide the parameters to determine the search space.

\section{Process and measurement noise estimation}\label{sec:q_and_r_estimation}

In \cite{Fucek2017}, the process noise was heuristically determined while the measurement noise was estimated from the cost function values of the neighboring disparities.
Since the reduction of disparity space search depends on the estimated covariance, which in turn is also a function of the aforementioned noises, it is important to know them as precisely as possible.
Therefore, we estimated the process and measurement noise from the ground truth data of the KITTI dataset that is obtained from the measurements of the 3D laser range sensor.
We extracted scenes that did not contain moving objects from 9 different sequences, which summed up to 7000 images in total.

\subsection{Process noise}

\begin{figure}
	\begin{center}
		\subfigure[Process noise]{\label{fig:process_noise_a}
    \includegraphics[width=0.38\columnwidth]{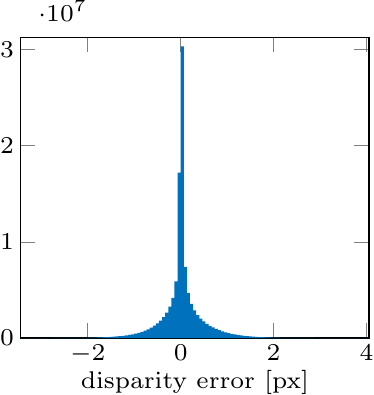}
    }
		\subfigure[Measurement noise]{\label{fig:measurement_noise_b}
    \includegraphics[width=0.38\columnwidth]{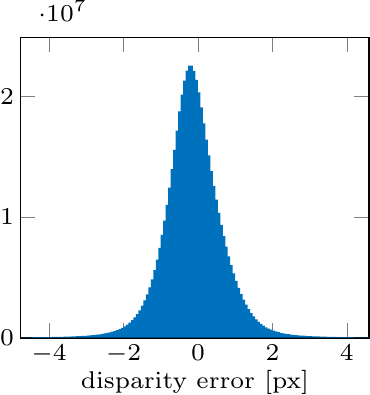}
    }
  \end{center}
		\caption{Distribution of error expressed in pixels. (a) process noise of ego-motion estimation and (b) measurement noise of proposed SGM implementation.}
		\label{fig:noise_histogram}
\end{figure}

\begin{figure}[!t]
	\begin{center}
		\subfigure{\includegraphics[width=0.38\textwidth]{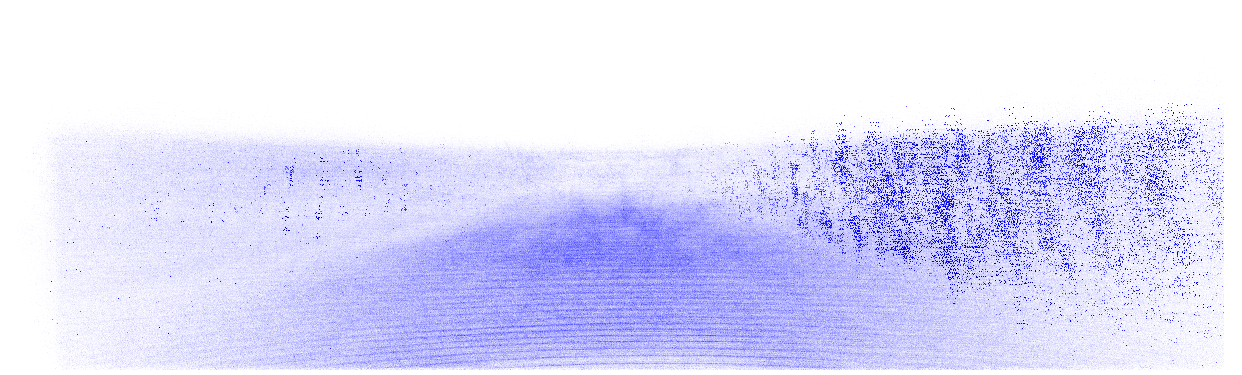}}
		\subfigure{\includegraphics[width=0.38\textwidth]{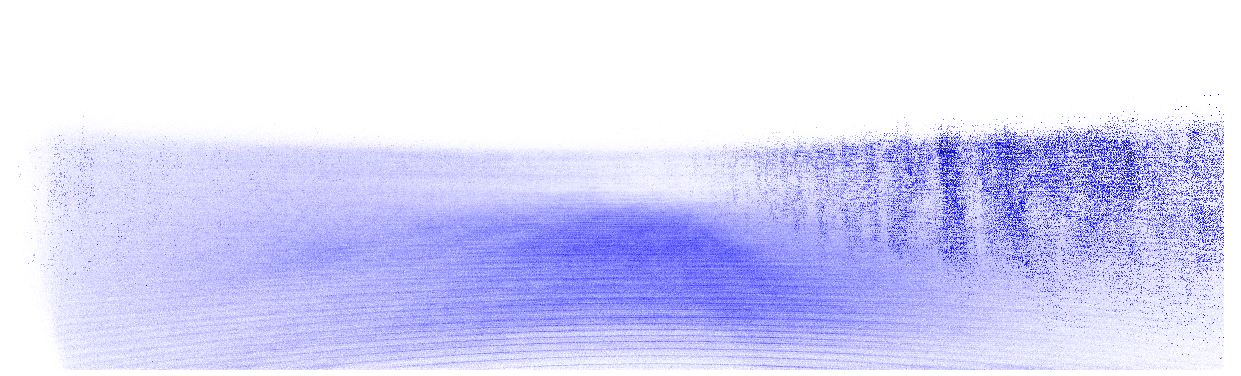}}
	\end{center}

		\caption{Accumulated absolute error of 2 different sequences. Intensity of blue color is proportional to the error.}
	\label{fig:accError}
\end{figure}

The source of the process noise is the uncertainty of ego-motion estimation.
To single out the error occurring due to ego-motion estimation, we used only the static parts of the sequences, without moving objects, where all the motion in the image comes from the motion of the camera.
Ego-motion was computed using the open source \texttt{libviso2} library \citep{Geiger2011}.
Each measurement was extracted using the intensity images of the stereo camera and depth information obtained by the 3D laser range sensor. 
The ego-motion algorithm estimated the transformation matrix using two consecutive stereo image pairs from the frames $k-1$ and $k$.
With the computed transformation matrix, we transformed the ground truth from $k-1$ to estimate the ground truth in $k$.
We assumed that the difference between the estimated disparity and ground truth in $k$ comes form the errors in the transformation matrix, and try to measure the variance of that error in the disparity space.
Figure~\ref{fig:process_noise_a} shows the histogram of the disparity error in pixels after applying estimated  ego-motion, from which we can see that more than $99\%$ errors are within $\pm 1$ pixel.
Figure~\ref{fig:accError} shows the result of error accumulated over two consecutive sequences.

\subsection{Measurement noise}
To estimate the measurement noise, we measured the error introduced by our implementation of the SGM algorithm.
Again, as in the case of process noise estimation, only scenes with static objects are used.
We used left and right intensity images and the 3D laser range sensor measurements in the single moment $k$.
Even though the measurements are conducted on image sequences for each image pair, the disparity is computed without reducing the disparity search space.
In other words, each image pair was processed as if it appeared at the initialization of the Kalman filter.
This way we avoid possible errors due to the reduced space and measure only the uncertainty of the SGM computation process.
The computed disparity between the left and right intensity image is compared to the measurements from the laser sensor and the difference is used for noise estimation.
Figure~\ref{fig:measurement_noise_b} shows the measurements of error values on one of the image sequences.

\section{Moving object detection}\label{sec:mov_obj_detection}

The estimated ego-motion can be used to predict correctly the disparity in the future frame for predominantly static scenes; naturally, up to the point of the estimated relative displacement accuracy and the amount of introduced novel  structure in the scene.
Since in the present paper we are not tracking dynamic objects through time, their displacement as well cannot be accounted for; therefore, the predicted and measured disparity maps will differ in the areas where moving objects are present.
Hence, moving objects are detected in the disparity domain without the need to take the intensity into account.
The goal is to obtain the bounding boxes surrounding moving objects by detecting the areas that contain high number of pixels whose predicted disparity differs from the measured one.

Moving objects are expected to be located in the parts of the image where the sum of the corresponding values contained in the submatrix containing absolute differences between the predicted and the measured disparities is high relative to its area.
The sum of all elements of a submatrix can be calculated in constant time using the \textit{inclusion-exclusion principle} with additional preprocessing.
First, small fragments of the image containing high amounts of disparity differences are detected using the sliding window approach.
Window sizes used in this step range from $20 \times 20$ to $50 \times 75$ pixels.
Sliding windows are moved only through the bottom two thirds of the image, thus avoiding excess computation as the upper third of the images in the KITTI dataset does not contain moving objects.
Due to using the aforementioned window sizes, detected areas of movement intersect and their sizes do not correspond to the exact sizes of the actual moving objects.
Those bounding boxes need to be processed further to obtain the wanted result, one bounding box per object, and they are grouped using a greedy approach.
In each iteration, a pair of boxes whose area of intersection relative to the area of their union is the highest is replaced by the smallest possible bounding box containing both of them.
As the final step, the bounding boxes of too small size are discarded, as they are unlikely to contain moving objects detectable by using the described method.

\begin{figure}[!t] \label{fig:detection}
	\begin{center}
		\subfigure{\includegraphics[width=0.4\textwidth]{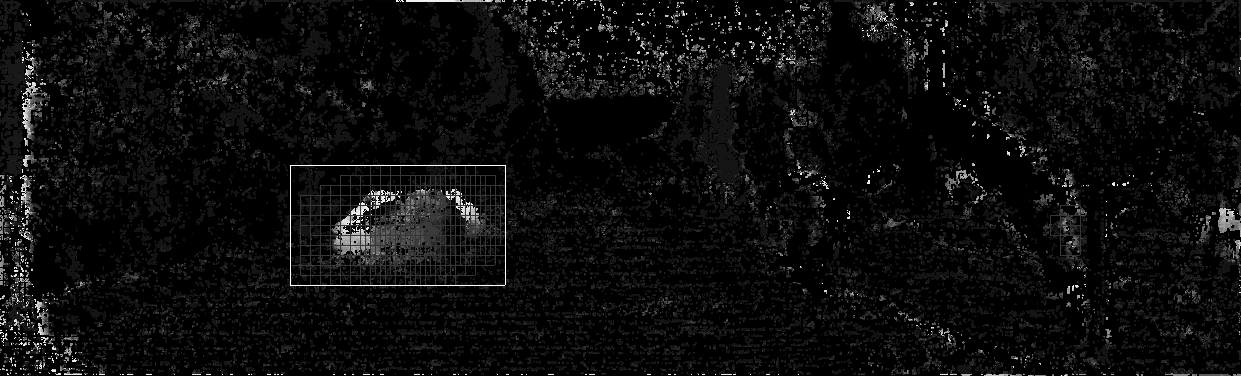}}
		\subfigure{\includegraphics[width=0.4\textwidth]{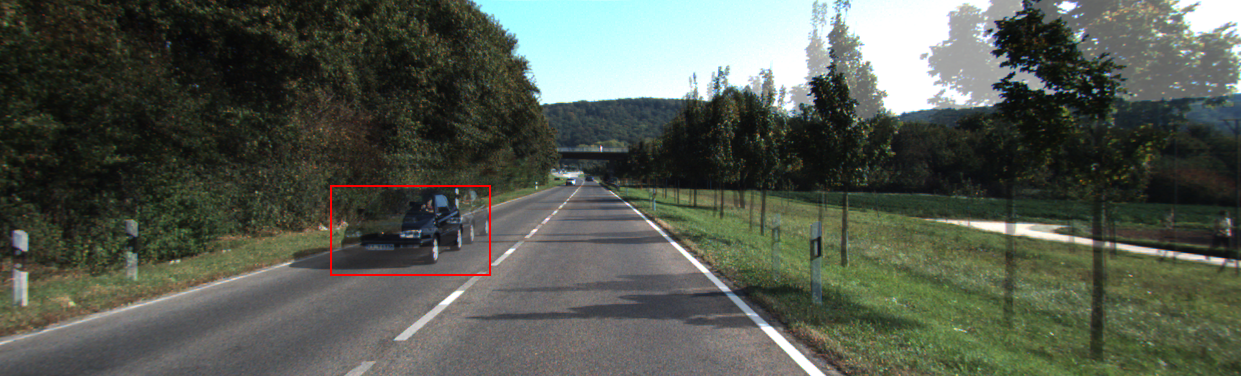}}
		\subfigure{\includegraphics[width=0.4\textwidth]{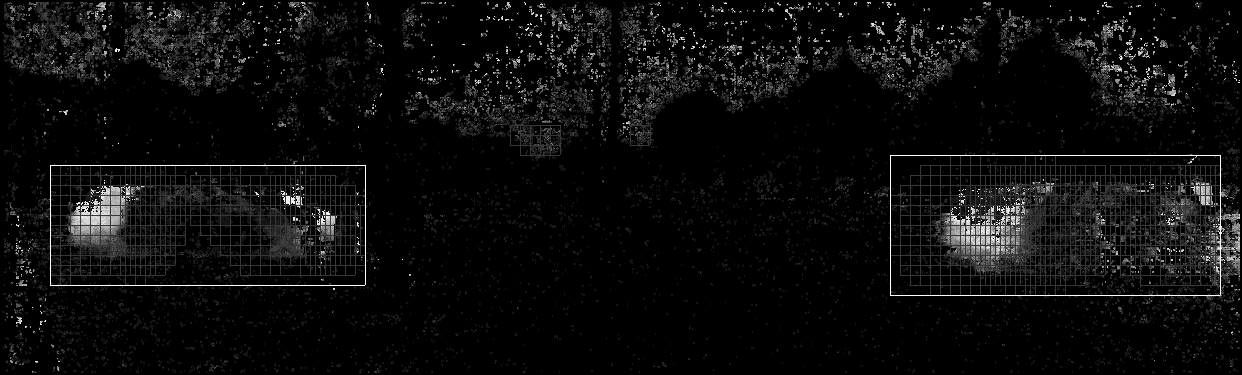}}
		\subfigure{\includegraphics[width=0.4\textwidth]{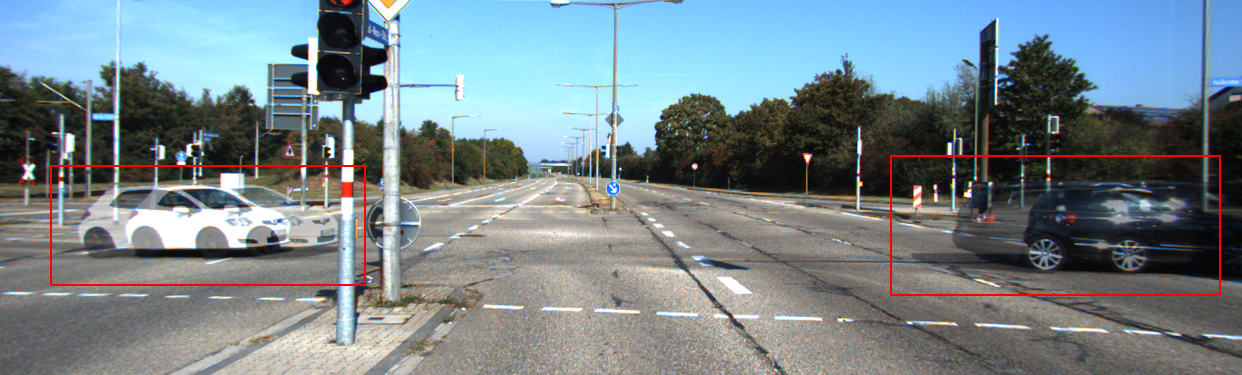}}
	\end{center}
		\caption{Moving object detection on two consecutive frames in scenes 183 and 45 from the \textit{data scene flow} set of images. The first and the third image show the difference between the predicted and the matched disparity maps for the respective scenes.}
\end{figure}

Moving objects in the KITTI dataset are mostly cars, cyclists and pedestrians.
With respect to their pattern of movement, they can be divided into two `base' groups: objects moving directly towards and away from the camera and objects moving perpendicular to the direction of camera movement.
When the moving object displacement is small, detecting motion for the former group can be challenging since it occurs directly along the projection line of the cameras, while for the latter group (especially large objects) it is possible that there will be no difference in the predicted and the measured disparity in the central area of the object as it does not change the depth of the scene.
Nevertheless, these are border cases and throughout the frames most objects exhibit combination of both motions.
In the top two images of Fig.~\ref{fig:detection} we can see a snapshot of detecting an incoming car.
It is important to point out that the presented method detects motion through two consecutive frames, as illustrated in the images in Fig.~\ref{fig:detection}, thus making the bounding boxes larger than the actual moving objects in some cases.

\section{Evaluation and results}
\label{sec:evaluation}
We evaluate the proposed algorithm on the sequences of the KITTI dataset.
Due to specific requirements of our algorithm, it was not possible to evaluate it on any of the official benchmarks.
The closest benchmarks are the \textit{sceneflow} benchmark and the \textit{depth} benchmark.
For the evaluation on the \textit{sceneflow} benchmark, we would also need the optical flow since the disparity ground truth is given only for the first frame.
On the other hand, the \textit{depth} benchmark only has one image per scene and we would not be able to show how the algorithm works in the reduced space.
Thus we used the raw data sequences that provide the ground truth (i.e. laser measurements) for multiple images per scene and for each moment ground truth was in the reference frame of the left image.
The implementation of the proposed algorithm is compared to the open source SGM implementation from the OpenCV library.

Before the evaluation, we used parts of the scenes without moving objects for tuning the parameters of the algorithms, namely, the the smoothness factors $P1$ and $P2$.
The parameters of the proposed algorithm were determined without reducing the disparity search space and without the Kalman filter.
The parametrization of OpenCV's SGM implementation was also conducted on the same sequences, and we tuned the parameters $P1$, $P2$ and the size of the window for block matching \texttt{SADWindowSize}, while other parameters were set to default values.
The parameters used in the evaluation are shown in Table~\ref{tab:parameters}.

\begin{table}
\centering
\caption{Parameters set in the evaluation}
	\label{tab:parameters}
    \begin{tabular}{ccc}
        Parameters         & OpenCV & proposed	\\
        $P_1$               & 26  & 6\\
		$P_2$				& 470 & 65\\
        \texttt{SADWindowSize}    & $3 \times 3$ & -\\
        \texttt{CensusWindowSize} & - & $5 \times 5$\\
        Loss acc. paths    & 8      & 4 \& 8\\
        Max. disp.         & 128    & 128\\
    \end{tabular}
\end{table}

\begin{figure}
  \centering
	\includegraphics[width=0.3785\textwidth]{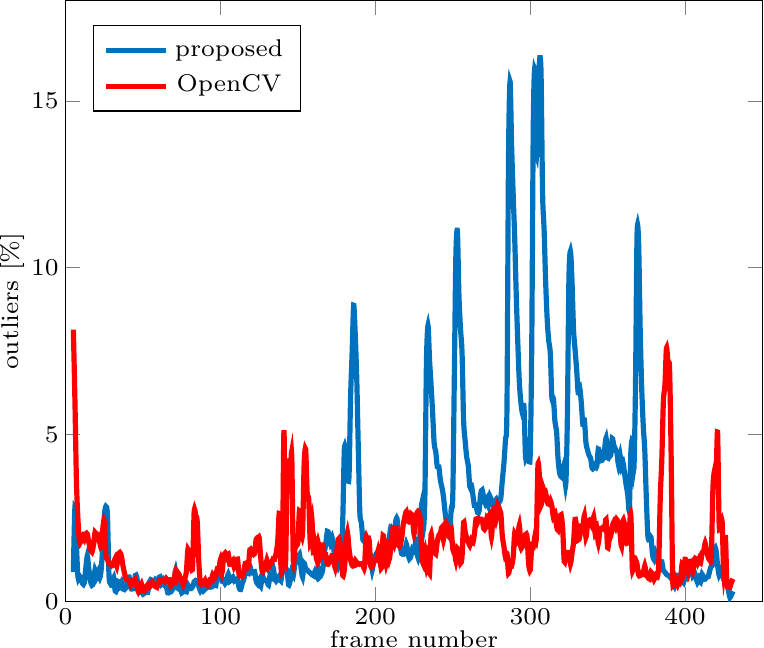}
    \label{fig:process_noise}
    \caption{Comparison of outliers computed with OpenCV SGM and the proposed approach in the sequence \textit{0051}\textsuperscript{\ref{fn:1}}. The peaks of the blue curve correspond to the frames with moving cars.}
    \vspace*{0.12cm}
\end{figure}

Evaluation was conducted on 7 scenes which were not used for the parameter tuning and noise estimation.
The results for each sequence are given in Table~\ref{tab:results_for_scenes}.
We can see that in all of the sequences, the average time needed to compute the disparity is smaller for the proposed algorithm than for the OpenCV's implementation.
It should be noted that the proposed algorithm, besides computing disparity using reduced search space, also runs disparity ego-motion based prediction and pixel-wise Kalman filtering.
In terms of accuracy, our implementation had on average fewer outliers in 6 out of 7 sequences.
In the problematic sequence, Figure~\ref{fig:process_noise}, the parts of the scene with a lot of moving objects have the highest impact on the error.

Since the number of accumulation paths in our implementation is a variable, in Table~\ref{tab:results_for_scenes} we also provide the execution time and accuracy for our implementation when only 4 accumulation paths are used.
As expected, the execution time is slightly shorter, but contrary to our expectation the accuracy is better when nondiagonal paths are used.
Authors in \cite{Spangenberg2013} explain that due to enviroment's structure, horizontal and vertical accumulation paths are more suitable for disparity computation.
By computing the disparity with 4 diagonal paths we noticed an increased number of outliers as shown in Table~\ref{tab:results_for_scenes}.
This explains better performance of nondiagonal 4-path implementation since the 8-path implementation takes also the errors of diagonal paths.

\begin{savenotes}
\begin{table*}[!t]
\centering
\caption{OpenCV and proposed implementation comparison. Outliers are defined as on the KITTI benchmark (absolute threshold of 3 pixels and relative threshold of 5$\%$). Diagonal and nondiagonal columns show the results computed using only 4 of 8 accumulations paths.}
	\label{tab:results_for_scenes}
    \begin{tabular}{|c|ccc|cccc|c|}
    		\multirow{4}{*}{Sequence} & \multicolumn{3}{|c|}{Time per image [s]}    			   &  \multicolumn{4}{|c|}{Outliers [\%]}  				   & \multirow{4}{*}{No. of images}\\
        						      & 	OpenCV		   & \multicolumn{2}{c|}{proposed}    &    OpenCV  & \multicolumn{3}{c|}{proposed} & \\
                       			      & \multirow{2}{*}{8-path}	& \multirow{2}{*}{4-path}  & \multirow{2}{*}{8-path} &  \multirow{2}{*}{8-path}	& \multicolumn{2}{c}{4-path} & \multirow{2}{*}{8-path} &\\
                       			      &	&	&	&	& nondiag.	& diag.	& &\\
        \hline
         \textsuperscript{\ref{fn:1}}\textit{19} & 0.25 & 0.17  & 0.19 & 1.07 & 0.58 & 2.11 & 0.83 & 395\\
         \textsuperscript{\ref{fn:1}}\textit{39} & 0.26 & 0.17 & 0.19 & 1.33 & 1.09 & 1.85 & 0.97 & 384\\
         \textsuperscript{\ref{fn:1}}\textit{51} & 0.28 & 0.16 & 0.18 & 1.72 & 2.41 & 4.49 & 2.72 & 427\\
         \textsuperscript{\ref{fn:1}}\textit{61} & 0.28 & 0.18 & 0.19 & 4.64 & 1.96 & 3.74 & 2.11 & 691\\
         \textsuperscript{\ref{fn:1}}\textit{84} & 0.26 & 0.16 & 0.18 & 1.27 & 0.75 & 1.70 & 0.76 & 372\\
         \textsuperscript{\ref{fn:1}}\textit{96} & 0.26 & 0.17 & 0.19 & 2.38 & 1.35 & 4.18 & 1.65 & 464\\
         \textsuperscript{\ref{fn:1}}\textit{117} & 0.26 & 0.17 & 0.19 & 2.93 & 1.54 & 3.88 & 1.75 & 649\\
    \end{tabular}
\end{table*}
\end{savenotes}

\footnotetext[1]{Full names of the sequences are: \texttt{2011\_09\_26\_drive\_\{0019, 0039, 0051, 0061, 0084, 0096, 0117\}\_sync} \label{fn:1}}

\section{Conclusion}\label{sec:conclusion}

In this paper we have presented how semi-global matching can be improved in terms of both speed and accuracy by using information from the previous frames, while also detecting moving objects in the scenes.
By using the ego-motion estimation and the disparity map of the previous frame, we can predict the disparity of the current frame and use it to reduce the complexity of stereo matching.
We use the Kalman filter to fuse the predicted disparity map and the reduced space semi-global matching measurement.
The predicted and the measured disparity will differ in the areas containing moving objects since their location cannot be accounted for using only this method.
Thus, we show how moving objects can be detected in the disparity domain without the need to consider the intensity images.
Furthermore, we estimate the process and measurement noise by conducting extensive experiments on the KITTI suite, on which we also validated our implementation of the reduced space semi-global matching.
Compared to the OpenCV SGM implementation, our method yields better results on the KITTI suite in terms of both speed and disparity accuracy.

\balance
\bibliography{ifacconf}             

\end{document}